# Deep learning and sub-word-unit approach in written art generation.

Krzysztof Wołk [1], Emilia Zawadzka-Gosk[1], Wojciech Czarnowski[2]

[1] Polish-Japanese Academy of Information Technology, Warsaw, Poland
[2] Jatar, Koszalin, Poland

{kwolk, ezawadzka}@pja.edu.pl, wcz@jatar.com.pl

**Abstract.** Automatic poetry generation is novel and interesting application of natural language processing research. It became more popular during the last few years due to the rapid development of technology and neural computing power. This line of research can be applied to the study of linguistics and literature, for social science experiments, or simply for entertainment. The most effective known method of artificial poem generation uses recurrent neural networks (RNN). We also used RNNs to generate poems in the style of Adam Mickiewicz. Our network was trained on the 'Sir Thaddeus' poem. For data pre-processing, we used a specialized stemming tool, which is one of the major innovations and contributions of this work. Our experiment was conducted on the source text, divided into sub-word units (at a level of resolution close to syllables). This approach is novel and is not often employed in the published literature. The sub-words units seem to be a natural choice for analysis of the Polish language, as the language is morphologically rich due to cases, gender forms and a large vocabulary. Moreover, 'Sir Thaddeus' contains rhymes, so the analysis of syllables can be meaningful. We verified our model with different settings for the temperature parameter, which controls the randomness of the generated text. We also compared our results with similar models trained on the same text but divided into characters (which is the most common approach alongside the use of full word units). The differences were tremendous. Our solution generated much better poems that were able to follow the metre and vocabulary of the source data text.

## 1  Introduction

Creativity is a magnificent human ability. Domains such as art, music and literature are thought of as requiring great creativity. One of the most demanding is poetry, as it needs to satisfy at least two prerequisites: content and aesthetics. Creating high quality poetry is thus demanding even for humans, and, until recently seemed almost impossible for machines. Until the 90s computer generated poetry was mostly restricted to science fiction such as Stanislaw Lem's 'Fables for Robots'.

Initial attempts of automatic poem creation were made in late 90s, and systems such as Hisar Manurung's application generating natural language strings with a rhythmic pattern were introduced. [6]

Automatic poetry generation was practiced as an art [5], hobby or entertainment, but also became a discipline of science, and an area of research in the natural language processing domain.

According to [7], approaches to poetry generation can be classified into four groups. Template based poetry generation, as in the ALAMO group, where the vocabulary of one poem and the structure from another can be used to create a new one. The Generate and Test approach, where WASP is an example. These types of systems produce random word sequences, which meet formal requirements, but are semantically challenged. There is also a group of applications with an evolutionary approach. One such example in this category is Levy's computer poet implemented as a neural network trained on data acquired from human testers. The fourth category includes systems that use a case-based reasoning approach. This strategy retrieves existing poems and adapts them to information provided by the user. One example in this category is the ASPID system.

Nowadays the most broadly used techniques for generating poems are neural-networks-based [3][4]. Different proposals for the language unit are considered [8]. Words [2] [10] are the most common element of learning and creating process, but there are also many systems based on characters or even phonemes [11]. In our work we introduce a new approach, using sub-word units for our implementation.

## 2  Experimental environment

We performed our experiment using the Google Collaboratory environment - a platform provided by Google for machine learning systems where Nvidia K80 graphic cards are available, and developed our models in Python with the PyTorch machine learning library.

## 3  Data pre-processing

In the first step, we loaded and pre-processed the corpus data. Uppercase and capitalized words needed to be annotated (during text tokenization) with respectively _up_ and _cap_ tokens. After tokenization the entire training text was lowercased. Our specialized stemmer program was used to divide the corpus into sub-word units [14], the tool was implemented by us especially for such purposes (it supports only Polish language). The program is a proprietary solution performing multiple operations to segment Polish texts into suffix, prefix, core and syllables. Moreover, it annotates texts with grammatical groups and tags it with part of speech (POS) tags. We used the application to split text in Polish into sub-word units. The program was run with the '7683' option, a combination of stemming and division options. According to the application documentation:

7683=1+2+512+1024+2048+4096 (divide (more) words with dictionary, then algorithmically and unknown words), where: 1 - stem "nie" prefix and 2 - stem extra prefixes are stemming options, 512 - divide words, 1024 - divide with dictionary, 2048 - divide algorithmically, 4096 - divide unknown words that belong to division options category.

In the next step we tokenized our corpus and created a list of all tokens. Additionally, we changed end of line characters into _eol_ token. For the syllables that do not exist in a corpus the _unk_ token was created. For remaining syllables and tokens, we added '++', and '--' as connection symbols (e.g. in '--PO++', the '_' tokens were removed).

Two auxiliary functions were also created, one transforming text into the list of created tokens and the other translating the list of tokens into text.

To prepare input data for the neural network (NN), we divided our data string into chunks of 400 syllables. Each chunk was converted into a LongTensor, by assigning an index from the _all_tokens list to each syllable. This pre-processing allowed us to create a set of input and target tensors.

## 4  RNN

In our experiment we used a Recurrent Neural Network. The RNN network construction enables the model to use the state from the previous step as its input information for the next step. This characteristic enables the network to learn from its 'experience'. The data from the previous steps affects the next step's results. This attribute of RNNs facilitates the discovery of patterns from sequences provided. Our input data is a poem in Polish, so we can treat it as a string with repeatable patterns.

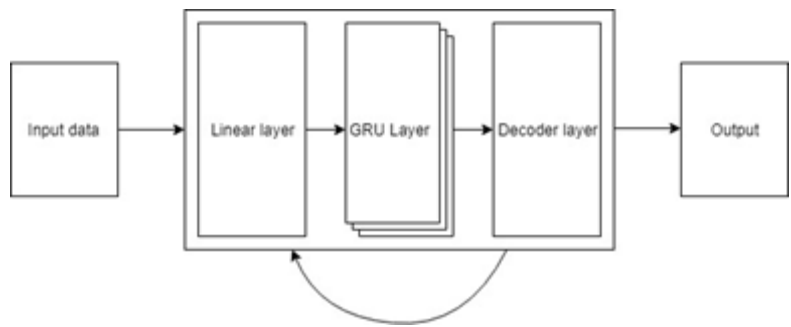

**Fig. 1.** Overview diagram of RNN used in experiment

Our RNN consists of three layers: a linear layer, one GRU layer and a final decoder layer. The linear layer codes the input signal. The GRU layer has multiple sub-layers, and operates on internal and hidden state events. The decoder layer returns a probability distribution. Our model received step t-1 token as its input with a target to output a token at step t.

## 5  Experiment

As the corpus for our experiment we used text of 'Pan Tadeusz' (Full title in Polish is: 'Pan Tadeusz, czyli ostatni zajazd na Litwie. Historia szlachecka z roku 1811 i 1812 we dwunastu księgach wierszem', in English: 'Sir Thaddeus, or the Last Lithuanian Foray: A Nobleman's Tale from the Years of 1811 and 1812 in Twelve Books of Verse'). It is an epic Polish poem written by Adam Mickiewicz [1]. The book was published in the 19th century, and consists of over 9000 verses and has Polish alexandrine metrical line, which means that every line has 13 syllables. The poem is regarded as a national epic of Poland and has been translated into many languages. Its first verses are as follows:

*KSIĘGA PIÉRWSZA.*
*GOSPODARSTWO.*
*TREŚĆ.*
*Powrot panicza -- Spotkanie się piérwsze w pokoiku, drugie u*
*stołu -- Ważna Sędziego nauka o grzeczności -- Podkomorzego uwagi*
*polityczne nad modami -- Początek sporu o Kusego i Sokoła -- Żale*
*Wojskiego -- Ostatni Woźny Trybunału -- Rzut oka na ówczesny stan*
*polityczny Litwy i Europy.*
*Litwo! Ojczyzno moja! ty jesteś jak zdrowie;*
*Ile cię trzeba cenić, ten tylko się dowie*
*Kto cię stracił. Dziś piękność twą w całéj ozdobie*
*Widzę i opisuję, bo tęsknię po tobie.*

In our experiment we loaded the corpus (as a big text file), tokenized it and divided it into syllables (sub-word units) with the '7683' option and our stemming procedure. Afterwards, an 'all_tokens' list was created with 5,059 different tokens found in our corpus. Pre-processed text was divided into chunks of 400 syllables. As the result we got 198 chunks (whole text was divided into 79,544 tokens).

To evaluate our network, we input one token at a time, use the outputs as a probability distribution for the next sub-word unit and repeat the action. To start the text generation, we directed the initial series to build the hidden state and generate one token at a time.

We used the following parameters to train our network:
Number of epochs = 15
Hidden layer size = 500
Number of layers = 3.

Additionally, we created a set of functions to monitor the training process. The loss function was used to monitor the training progress, while training time was also monitored. The 15 epochs of training took 96 minutes using Google Collaboratory platform.

The loss function is an indicator describing a dissimilarity between real and desired output of neural network. Lower values of loss suggest better accuracy for output values. Figure 2 demonstrates values of the loss during the whole training of our network. The system is adapting its parameters into expected outcome with every iteration of the training, therefore it can be observed a decreasing tendency of the loss changes, with the final value below 1.

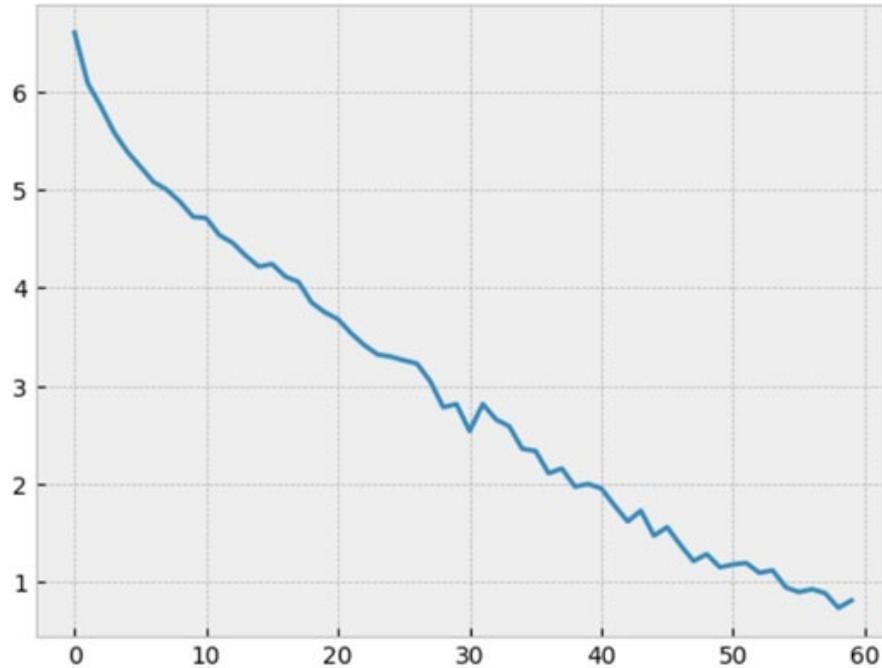

**Fig. 2.** Loss values during network training

Our trained network generated this poem in the style of Adam Mickiewicz:

*Litwo! Ojczyzno moja! czaśłam w Wilnie,*
*Lecz niedźwiedź w pole; więc Sędzia z Domcom,*
*Niech co to nie wiedziacie rzegnąć swą rómą,*
*Szlachta odwieczna, w któréj otwierzycie?*
*Czy to Rejent równie podszepnął: już do polu żeca,*
*Pokojowa zaś znalazł się przeznało*
*Odbijamy od drzewo Rejent, zawołał Rejent,*
*Od Bonaparte znowu w kota się przerzuca,*
*Daléj drzeć pazurami, a Suwarów w kuca.*
*Obaczcież co się stało w milczeniu głębokiém.*
*Sywciąż wtenczas widząc, jak od chmielu tyki,*
*W kurtkach, w budowne jak chartaści i urzęwe dawał.*

This text was written with a metre similar to Polish alexandrine. The generated verses length is mostly thirteen syllables, but does not always meets this criterium. We also can observe proper punctuation and text capitalization. The vocabulary seems to also be similar to Mickiewicz's. It can be seen that although the whole creation does not have an overall coherent theme or clear semantic structure, some of the individual verses do create meaningful sentences.

Another important indicator we decided to monitor during network training was 'bad-words ratio'. This metric informed us about the number of syllables not used for word creation. Figure 3 presents values of this measure over time. Ratio of the syllables not used for words formation is quite low after the training. It decreased (from the first iteration) from almost 30% to less than 2% after 5 epochs. It signifies that our network learned how to use the syllables provided in the inputted data. It also proves that the sub-word approach works as we anticipated.

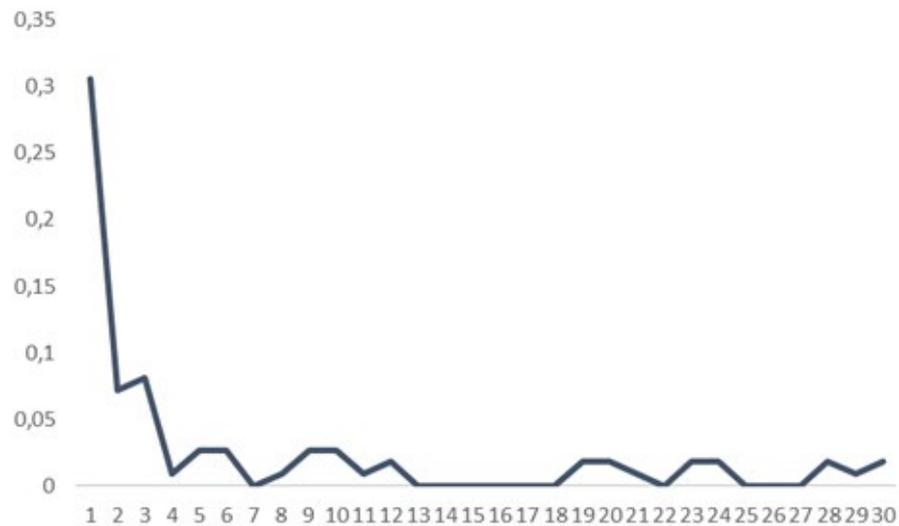

**Fig. 3.** Bad-words ratio values during network training

## 6 Results

To control the randomness of the generated poem we used a temperature hyperparameter. Depending on the temperature value, different and interesting results are generated. Lower values make the output data more probable given the training data. Values around 1 or higher make the outputs more creative, whereas values closer to 0 force the network to generate only the most probable results.

To be more precise, the temperature is a hyperparameter of LSTM's (and neural networks generally) used to control the randomness of predictions by scaling the logits before applying softmax. For example, in TensorFlow's Magenta implementation of LSTMs, temperature represents how much to divide the logits by before computing the softmax.

When the temperature is 1, we compute the softmax directly on the logits (the unscaled output of earlier layers) and using a temperature of 0.6 the model computes the softmax on logits/0.6, resulting in a larger value. Performing softmax on larger values makes the LSTM more confident (less input is needed to activate the output layer) but also more conservative in its samples (it is less likely to sample from unlikely candidates). Using a higher temperature produces a softer probability distribution over the classes and makes the RNN more "easily excited" by samples, resulting in more diversity and more mistakes as well.

Neural networks produce class probabilities with logit vector z where z=z=(z1,…,zn) by performing the softmax function to produce probability vector q=(q1,…,qn) by comparing zi with with the other logits.

$$q_i = \frac{\exp(z_i/T)}{\sum_j \exp(z_j/T)}$$

is the temperature parameter, normally set to 1.

The softmax function normalizes the candidates at each iteration of the network based on their exponential values by ensuring the network outputs are all between zero and one at every timestep. Temperature therefore increases the sensitivity to low probability candidates. [16]

The effects were surprising and interesting. Setting the temperature to a value of 0.8, the network generated the following poem, the first lines of which are presented below.

*Litwo! Ojczyzno moja!*
*To za Dowejce i stanęli Mateczniku,*
*Wieczają tylko polski Tadeuszkę ożeką,*
*I zamiast iść do Panów, hałasy że cieli*
*Toczyli w krzesłach w powiewawych zadartym, świecie bez rury,*
*Spowiedzi się; Półkozic jaśniał w las,*
*A Gerwazy; Skołuba z zapaści przymierzy,*
*Czyli las był; z ich wszystkich zaoczyny.*
*Przerwał je stare, albo raz niewiewał uczyć,*
*Niby krano mumienne po trawie, robacze,*
*Marciały, odgadniesz, że na charty po chwili.*
*Czyli też zdarzeczka, czyli widzisz, kieli*

The result seems to be diverse. Most of the verses and sentences does not have any sense, although the majority of the words are understandable and exist in Polish vocabulary. Metre is close to Polish alexandrine.

Subsequently, when the parameter was changed to 0.2 the following was obtained.

*Litwo! Ojczyzno moja!*
*pradzi Jeden wielki myślili -- Cham, Skołuba krzyknęła*
*czówna, ogon nim do mnie się zabawić! -- Ja myśliwiec!*
*przerwałem (tu do wielkiéj pozwory? Doweyko!*
*nierwała Tekomości! Podczaszyc na wielkim gniewie, Ale Hrabia,*
*A Pan Hrabię i Sędzia tuż przy Berdziego Boże. Ja, Sędziecznym młody, pytany o zdromie*
*polityczny Litwie monarcha myślani*
*młowłaśni młodzi rzecz spadał towarzyszy,*
*przerwała Tekomości! to oni!*

*przerwałem że czasem wystrzeliła*

The Poem generated with temperature=0.2 consists of more real Polish words and even whole phrases. The metre seems to be more distant from our training data, but still imitates Polish alexandrine.

Selecting higher temperature values results in more random output values. For experimental purposes we decided to apply a temperature as high as 1.4. The resulting poem is presented below.

*Litwo! Ojczyzno moja! ty wszystko, usłynił kitdą zostaną, zda ciedzenie przestrzyni*
*Leste, tém zrarza ku otbią mami rzędawał,*
*Niby wódz widziałam była po leśka zwierz siedzenie,*
*Chce je trzewałem ska wielka, będzie o otwiemie!*
*Bo deusz na może świępać i siebie po dworze*
*Jéj żeby się pokładał obławą i świamę.*
*Wstępali prozem o malarstwie w komciadki,*
*Zosia w lot przywiedzenie się milczą się gniewa,*
*Tam dęsałam i ze czuły: pas konato,*
*Pilnami mu po drzeni, owchalby charty,*
*Szła mamą! cóż sach tysiąc złota już grunt dawem,*
*Niby Pan krzywał się, lecz*

The resultant text consists of valid Polish words, but also a number of invalid expressions. No consistent meaning can be found in the generated text, although it still mostly follows the desired metre.

As our approach to poetry generation is quite novel, we decided to juxtapose our results with text generated character by character by a recurrent neural network. The network was trained with following parameters:

n_epochs = 50
hidden_size = 400
n_layers = 2

The number of training epochs was increased to 50, but the network consisted of 2 layers. The training time was similar in both experiments and took about 1.5 hours.

After pre-processing the necessary data and training the network we generated poems using three different temperatures. Following previous experiments, the temperature indicator was set to 0.8 with the following results presented below.

*Pana,*
*Pajał wim storemymną, tery tejy;*
*Wychynią szwogole, jać jaka,*
*Skrzekło udy, wniekoła,*
*I nudość wisi mała wieryży słajwie wła udawytył oporaźny,*
*Sona, świ, przykił szeną roż stestwe wydowniewa neny,*

The poem generated by the char-based network does not follow the Polish alexandrine metre. Created words are stylized into the poet's vocabulary, but most of them do not qualify as correct words in Polish language. The poem itself cannot be classified as meaningful.

As lower temperatures enable the generation of more probable output, we present below a poem created using temperature 0.2.

*Thy, odo się się wie się się szada wi się wie się się wie szała wie się wie wie sza się wie się wie się szała, wie wyła, wie wie do wie się się wie podała, wie się się się wie do wie jak wie szała w*

The generated text does not look like a poem at all. Words created by the NN are real Polish expressions, but they do not have any schema or phrasal meaning.

The higher temperatures produce more random results. The last example was created with temperature=1.4.

*Thecegéj utéjégła! nguscy, ażąc. VąrŚni.ż dŻóc nię; wą Dolu chogyśsakiękdy uwy; o_jicpaże, Ban desie je każymik!*
*Akyski nie żessyl kórzcierbłéj awwi,, hrzeliwachrzeguj.*
*A rwum mawlakguł łyłrąy, arzymu*

The text looks very incoherent. Almost all created strings do not belong to Polish vocabulary. No metre can be noticed.

Additionally, it is worth paying attention to the influence of the sequence length on the recognition of distant relationships. 30-50 tokens is a window in which the network remembers the sequence structure well. Using syllables instead of characters shortens (compresses) the sequence approximately 3 times, which makes training easier. The network can 'analyze' the structure of several rows at once. Potentially using LSTM instead of GRU can give even better results when modeling long-distance dependencies.

## 7 Discussion and conclusion

Polish is a complex language with complicated grammar and vocabulary. The language has three tenses. Verbs are in perfect or imperfect form, also they differ for gender and number. There are seven cases of a noun: nominative, genitive, dative, accusative, instrumental, locative, vocative. Nouns also occur in two numbers (singular and plural) and three gender forms (masculine, feminine and neuter). Adjectives also have gender, case and number. All those rules cause the complexity of language. The vocabulary (language corpus) is very large due to the morphological richness of the Polish language. Therefore, any automatic analysis or generation of Polish language is difficult. [12]

In this paper we proposed generating author-stylized poems using innovative sub-word units, based on the example of 'Sir Thaddeus'. Our approach focuses on sub-word units. In the solution we created, we tried to generate a poem congenial to Adam Mickiewicz's 'Sir Thaddeus' book. We used an RNN for our work. The neural network was trained on Mickiewicz's poem, based on a sub-word unit (close to syllable) corpus.

After preparation of the data with our special stemmer tool and the network training we compared the results of the generated text using three temperatures: 0.8, 0.2 and 1.4. All three cases were able to generate quite promising texts, imitating 'Sir Thaddeus'. The 0.2-temperature generated a poem with most understandable words and phrases, as we expected. In all three cases the metre was at least partially preserved. The general vocabulary was epitomizing Mickiewicz's.

For comparison, we conducted a similar experiment using a classical solution with characters as the unit of text generation. The differences between the results was tremendous. The neural network trained on the poem divided into characters, after 50 epochs of training demonstrated much worse results. The experiment performed with a temperature of 1.4 generated text resembling a random string, with no metre present. Setting temperature to 0.2, the generated text consisted of correct Polish words, but they did not construct any phrases and there was no metre. Only a temperature value equal to 0.8 created an impression of a poem, but its quality was much lower than that generated with sub-words units.

The poems generated by our solution cannot yet be confused with poems generated by humans, but they clearly recognize and follow stylistic and poetic rules applied by the author of the poem used as training data. The attempt to create the resultant text in Polish alexandrine is clearly visible in the shown examples. The word formation is quite good, the number of incorrect phrases is low. Punctuation and capitalization were satisfying. It was also the case that the method we used in our research provides much better results than when the unit of the training data is a character.

We plan to continue our research on automatic poetry generation using various training data. As units different to character or word gave promising results, we will experiment with various language units and the data amount of training data. Modifying the parameters of our network will also be considered in our future work. The temperature dimension might be adjusted more accurately. In our further work, we plan to perform fine-tuning of the hyperparameters of our neural network. We would like to use the Cyclical Learning Rates method [13] to adjust learning rate of our RNN. Finally, we plan to annotate our training texts with POS tags (and/or grammatical groups) and interpolate the results with the Byte Pair Encoding (BPE) algorithm [16].

Performing our research, we concluded that although generating poetry (or other texts) does not seem to be a pressing concern, it can bring many benefits to several domains. It could be broadly used in entertainment, to generate texts, poems or even whole books or book series. Discovering the pattern of a specific author's style might be used to adapt existing text to the specific language of the selected author, era or type. It can also be used to for author style transfer or style mixing. It allows us to construct any lyric in any style imitating any epoch. Such a solution might also be valuable education. Recognizing the patterns behind a specific author will be useful to discover all the books written by the same author but under different pseudonyms, e.g. Stephen King. This writer published as Richard Bachman, because the editor did not want to agree to publish more than one book a year [15]. Also, J. K. Rowling, the author of the 'Harry Potter 'series, after a successful saga about wizards, decided to publish as Robert Galbraith. With automatic style recognition it might be much easier to identify a writer. The idea could be especially useful for the authors living in previous ages, as usually the data available for researchers is very limited, so the authorship can be mainly confirmed by the analysis of the text. The main domain where our research can be applied is literature and linguistic. It can be imagined that generating author-stylized text might expand the field of literature analysis.

Finally, we see the potential improvement of quality in the use of a non-direct but general language model of the Polish, projected on the language model of poetry and ultimately projected on a specific author (e.g. "Pan Tadeusz"), completed with fine-tuning. [18] This will allow us to learn more about poetry and language structure.

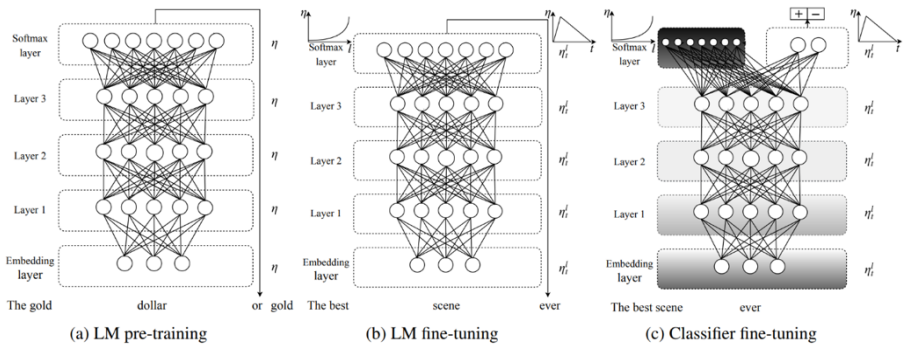

**Fig. 4.** Language model adaptation.

In such approach we can distinguish three stages: a) The LM is trained on a general-domain corpus to capture general features of the language in different layers, b) the full LM is fine-tuned on target task data using discriminative fine-tuning ('Discr') and slanted triangular learning rates (STLR) to learn task-specific features, c) the classifier is fine-tuned on the target task using gradual unfreezing, 'Discr', and STLR to preserve low-level representations and adapt high-level ones (shaded: unfreezing stages; black: frozen). [18]

We can also use Transformer architecture to better model distant relationships [17]. Additionally, natural language processing tasks, such as caption generation and machine translation, involve generating sequences of words. Models developed for these problems often operate by generating probability distributions across the vocabulary of output words and it is up to decoding algorithms to sample the probability distributions to generate the most likely sequences of words. For this problem we would like to use in future a popular heuristic, which is the beam search. It expands upon the most common greedy search and returns a list of most likely output sequences. Instead of greedily choosing the most likely next step as the sequence is constructed, the beam search expands all possible next steps and keeps the k most likely, where k is a user-specified parameter and controls the number of beams or parallel searches through the sequence of probabilities. This would most likely lead to better art generation as well. [19]

Analysing and recognizing text might be useful in forensics, in verification and evaluation of evidence of crime. Finally, social sciences and art can benefit from using artificially created texts.